\documentclass{article}

\usepackage[main, final]{neurips_2026}
\usepackage[utf8]{inputenc}
\usepackage[T1]{fontenc}
\usepackage{hyperref}
\usepackage{url}
\usepackage{booktabs}
\usepackage{amsfonts}
\usepackage{amssymb}
\usepackage{amsmath}
\usepackage{nicefrac}
\usepackage{microtype}
\usepackage{xcolor}
\usepackage{graphicx}
\usepackage{algorithm}
\usepackage{algorithmic}
\usepackage{capt-of}
\usepackage{wrapfig}

\title{RoSA: Rotational Sparse Adaptation for Memory-Efficient Fine-Tuning}

\author{Muhammad Azeem Lodhi$^{1}$ \quad Chao Zhou$^{2}$ \quad Rebekka Burkholz$^{2}$ \\
  $^{1}$Saarland University, Saarbr\"ucken, Germany \\
  $^{2}$CISPA Helmholtz Center for Information Security, Saarbr\"ucken, Germany \\
  \texttt{mulo00001@stud.uni-saarland.de}, \ \texttt{\{chao.zhou, burkholz\}@cispa.de}
}

\begin{document}

\maketitle

\begin{abstract}
Parameter-efficient fine-tuning (PEFT) reduces the cost of adapting foundation models by focusing training on a small parameter subset. Complementary to this idea, we introduce RoSA (Rotational Sparse Adaptation), which narrows adaptation to a subset of layers at a time. RoSA freezes lower layers close to the input throughout training and rotates a trainable block over later layers, progressively increasing the number of frozen layers close to the input. This design reduces optimizer-state memory, shortens backpropagation, and even forward propagation if activations at the last frozen layer are cached. Because RoSA is orthogonal to the choice of trainable parameterization, it can be combined with PEFT methods or sparse optimizers within each active block. Experiments across multiple LLM architectures and tasks show that RoSA reduces peak memory while maintaining strong fine-tuning performance.
\end{abstract}

\section{Introduction}

Fine-tuning Large Language Models (LLMs) has become one of the standard ways to adapt foundation models to downstream tasks. However, as model scale increases, the main bottleneck in training is often not the number of parameters alone, but the memory required to store optimizer states and activations. For a 7-billion-parameter model such as Llama-2 \citep{touvron2023llama}, model weights can fit comfortably for inference, but full fine-tuning with AdamW \citep{loshchilov2017decoupled} requires substantially more memory because optimizer states must be maintained for all trainable parameters. This makes dense fine-tuning difficult on commodity hardware and pushes many practitioners toward lighter adaptation methods.

Parameter-efficient fine-tuning (PEFT) methods address this problem by freezing most of the pretrained model and restricting learning to a small trainable subspace. Adapters \citep{houlsby2019parameter}, LoRA \citep{hu2021lora}, QLoRA \citep{dettmers2024qlora}, and DoRA \citep{liu2024dora} have shown that strong downstream performance is often possible with far fewer trainable parameters. These methods are practical and effective, but they reduce memory by limiting what can change during training. This creates a different trade-off: the model becomes easier to adapt under tight memory budgets, but the adaptation capacity is tied to a fixed low-rank or adapter-based parameterization.

In this work, we study a different way to reduce memory: instead of keeping the same small subset of parameters trainable throughout optimization, we ask whether the trainable part of the model can move over time while a large input-side prefix remains frozen. This leads to RoSA (\textit{Rotational Sparse Adaptation}), a fine-tuning framework that freezes the early layers, keeps an output-side anchor active, and rotates a trainable block through the later layers. The frozen prefix is not only a parameter-saving device. Since these layers do not receive updates, RoSA avoids storing optimizer states and gradients for them, and the backward pass does not need to continue through this prefix. In an optimized system implementation, the same fixed prefix could also be precomputed, cached, and compressed into reusable activations, reducing repeated forward computation;
our current implementation does not do this and still computes the full forward pass. At any training step, only one contiguous group of later layers is active together with the output-side anchor, so the instantaneous memory footprint remains small. Across training, different parts of the model are updated at different times.

RoSA can also be combined with sparse optimization inside the active block. In our default experimental instantiation, we use NanoAdam \citep{zhou2025pay}, which maintains optimizer states only for a selected subset of parameters within the active layers. This gives RoSA two separate mechanisms for reducing memory. Rotation limits how many layers are active at once, while sparse optimization limits how many parameters inside those layers require optimizer state. As a result, memory scales with the active part of the model rather than the full depth of the network.

\begin{figure*}[t]
    \centering
    \includegraphics[width=0.72\textwidth]{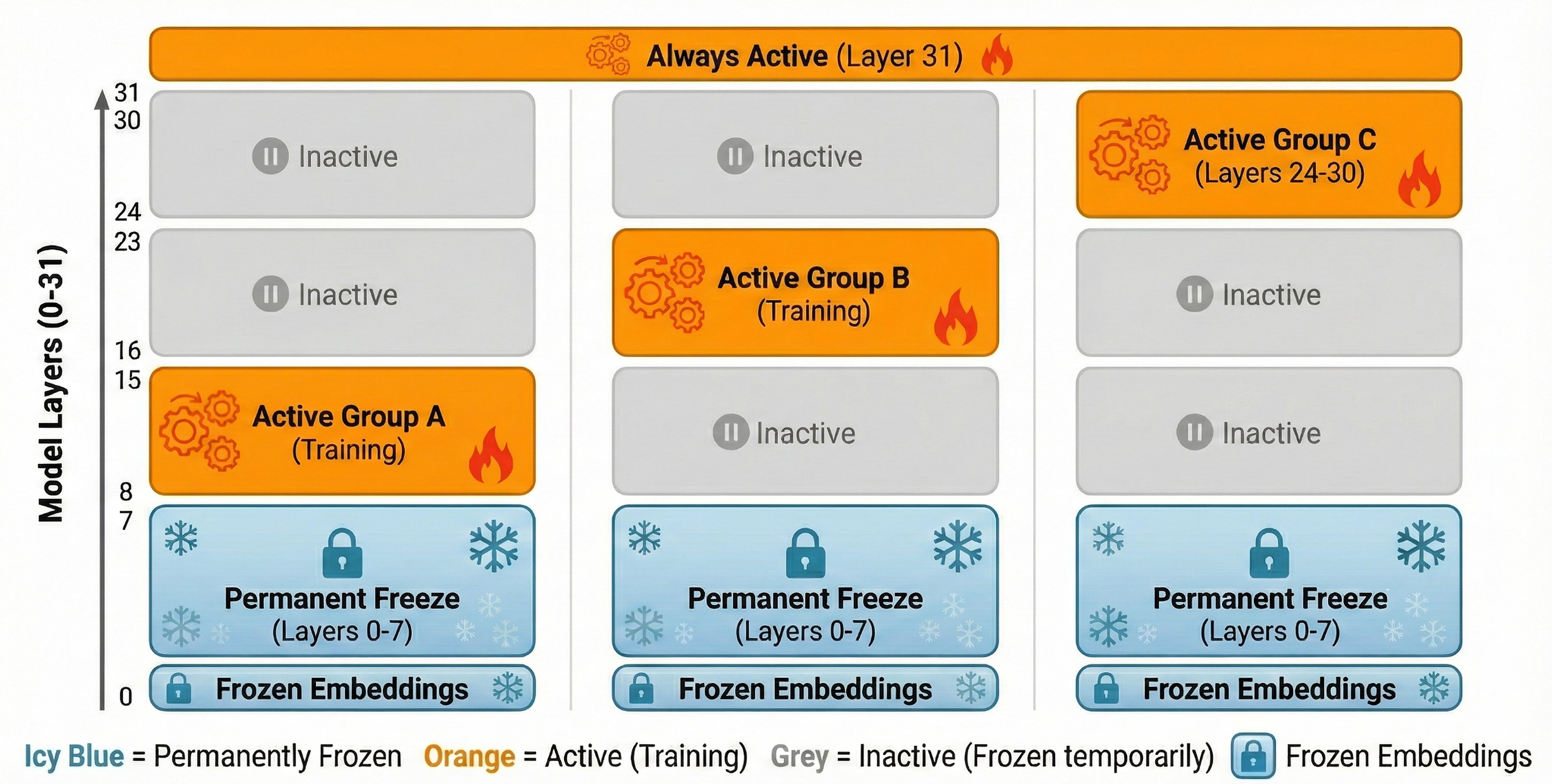}
   \caption{RoSA training schedule. RoSA freezes a large input-side prefix, keeps an output-side anchor active, and rotates one consecutive trainable group through the later layers. Freezing the prefix removes optimizer states and parameter gradients for early layers, shortens the backward path, and enables future activation caching for the fixed prefix.}
    \label{fig:rosa_schedule}
\end{figure*}

Our goal is not to present RoSA as a competitor or replacement for LoRA and related PEFT methods, but rather as a complementary framework that operates on an entirely orthogonal axis of the design space. Because RoSA regulates when layers are updated rather than how the trainable parameters are represented, it acts as a versatile memory-reduction mechanism. It can be applied to dense fine-tuning to make it feasible on commodity hardware, or it can be seamlessly composed with PEFT methods like DoRA to reduce their already small VRAM footprints even further. This structural perspective also separates RoSA from prior block-wise approaches such as BAdam \citep{chen2024badam}. While BAdam functions as a block-coordinate optimizer—reducing memory by updating one parameter partition at a time—RoSA is a depth-aware scheduling framework built around a permanently frozen prefix, a rotating active layer group, and an always-active output anchor. Consequently, RoSA not only limits optimizer-state memory but also actively exploits the sequential structure of Transformers to shorten the backward path as the active block moves deeper in the network.

We evaluate RoSA on Llama-2-7B and show that it reduces peak VRAM from 78.5\,GiB for full fine-tuning with AdamW to 34.9\,GiB while reaching 79.61\% average accuracy on eight commonsense reasoning benchmarks, close to the 79.99\% full fine-tuning baseline. We further show that rotation yields additional memory reduction beyond sparse optimization alone, that RoSA is not highly sensitive to the exact active-group size, and that the method transfers beyond the original setting to other models such as Qwen-2.5-3B and Mistral-v0.1-7B, and to other tasks such as GSM8K.
Rotating the active block also outperforms training a fixed subset of layers, even when the fixed subset contains twice as many layers, and at matched memory RoSA outperforms layer-wise importance sampling (LISA) by more than 8 points.
Together, these results suggest that rotational training is a simple and effective way to trade time for memory while preserving much of the benefit of dense adaptation.

In summary, this paper makes three contributions. First, we introduce RoSA, a rotational fine-tuning framework that freezes a large input-side prefix and moves the trainable layer block over time instead of fixing a small trainable subspace throughout training. Second, we show that this prefix-plus-rotation design reduces memory in multiple ways: it removes optimizer states for frozen layers, restricts optimizer states to the active block, and shortens the backward path as the active block moves deeper. Third, we empirically evaluate RoSA across memory, accuracy, and training-time trade-offs: it reaches near-full-fine-tuning accuracy on Llama-2-7B while fitting within a 40\,GB memory budget, remains stable across multiple rotation schedules and random seeds, transfers to Qwen-2.5-3B, Mistral-v0.1-7B, and GSM8K, composes with PEFT methods such as DoRA, and compares favorably with static layer subsets as well as with BAdam, LISA, and GaLore.

\subsection*{Related Work}

\textbf{Parameter-efficient fine-tuning.}
Parameter-efficient fine-tuning (PEFT) methods reduce memory by freezing the pretrained backbone and training only a small number of additional or reparameterized weights. Adapters \citep{houlsby2019parameter} introduce small trainable bottleneck modules inside the Transformer, while LoRA \citep{hu2021lora} updates low-rank matrices instead of dense weights. Later variants such as AdaLoRA \citep{zhang2023adalora}, QLoRA \citep{dettmers2024qlora}, and DoRA \citep{liu2024dora} improve this recipe through adaptive rank allocation, quantized backbones, or weight decomposition. These methods are highly effective when memory is the main constraint, but they reduce memory by restricting the trainable subspace itself. RoSA addresses a different part of the design space. Instead of fixing a small subspace throughout training, it changes which layers are trainable over time, so the instantaneous memory cost stays low while the cumulative set of updated parameters can be much larger.

\textbf{Memory-efficient dense optimization.}
A second line of work aims to preserve dense optimization while reducing optimizer or training memory. Quantized optimizers such as 8-bit Adam reduce state precision to lower memory overhead \citep{dettmers20228bit}. Gradient-based compression methods such as GaLore project updates into low-rank gradient subspaces while keeping full-parameter learning \citep{zhao2024galore}. Sparse or compressed adaptive optimizers such as MicroAdam \citep{modoranu2024microadam}, NanoAdam \citep{zhou2025pay}, and LOMO \citep{lv2023full} reduce the cost of storing or applying adaptive states under limited memory budgets.
Other systems-level approaches trade compute for memory by changing activation storage or layer execution, for example through reversible blocks as in RevFFN \citep{liu2025revffn}.
Furthermore, some optimizers have been designed specifically to sparsify neural networks during optimization \cite{gadhikar2024masks,jacobs2024maskmirrorimplicitsparsification,adnan2026sparseopt} or work well in the context of sparse training \citep{gadhikar2025signinlotteryreparameterizingsparse,jacobs2025hamhyperbolicstepregulate}.
RoSA is complementary to this literature. Its main contribution is not a new optimizer or compression rule, but a training schedule that reduces memory at the layer level. In our setting, sparse optimization is one component of the method, but rotation contributes an additional reduction beyond sparsity alone.
In the same way, low-rank gradient projection as in GaLore or quantized optimizer states could be applied inside the active block.

\textbf{Block-wise and layer-wise training.}
The idea of updating different parts of a network at different times has a long history. Greedy layer-wise training studied sequential optimization across depth in early deep networks \citep{bengio2006greedy}, and FreezeOut later showed that progressively freezing layers can reduce training time by shortening the backward pass \citep{brock2017freezeout}. Recent LLM fine-tuning methods revisit this idea from different angles. BAdam uses a block-coordinate Adam procedure, partitioning model parameters into blocks and updating one block at a time to reduce optimizer-state memory \citep{chen2024badam}. LISA applies layer-wise importance sampling and freezes many layers during optimization, showing that layer-wise non-uniformity can be exploited for memory-efficient fine-tuning \citep{pan2024lisa}. SALF studies semantic-aware layer freezing and emphasizes that choosing where to fine-tune can reduce backpropagation cost at the layer level \citep{gu2025salf}. RoSA is related to these methods, but differs in its depth-structured schedule: it freezes a consecutive input-side prefix, rotates a consecutive active block through the remaining layers, and keeps an output-side anchor active.
Concretely, FreezeOut starts from a fully trainable network, so its peak memory is that of full fine-tuning even though later iterations become cheaper. LISA samples its active layers at random, so the depth of the backward pass is not bounded by a fixed frozen prefix. BAdam cycles over blocks that cover the full depth and updates each active block densely, without a permanently active output-side block. Section~\ref{sec:experiments} compares RoSA empirically with BAdam, LISA, and static layer subsets.
RoSA's structure reduces optimizer-state memory while also shortening the backward path as the active block moves deeper. In future implementations, the fixed frozen prefix can also enable activation caching to avoid repeated forward computation through the same early layers. Thus, RoSA is not only a block-wise optimizer-state reduction method; it is a layer-scheduling framework that benefits from the sequential structure of Transformer depth.

\textbf{Layer roles, redundancy, and pruning in LLMs.}
Recent work has shown that different Transformer layers contribute differently to downstream behavior. Several studies report substantial depth redundancy and show that many layers can be pruned with limited loss on some tasks \citep{gromov2024unreasonable, men2024shortgpt, sun2025curse}. At the same time, more detailed analyses suggest that layer importance depends strongly on the task and evaluation protocol, with deeper or middle layers playing a larger role in reasoning than in retrieval-style behavior \citep{song2025demystifying}. RoSA is motivated by this heterogeneous view of depth. Rather than pruning layers permanently or assuming that all layers should always remain equally active, it uses a rotational schedule to allocate memory over depth during training. In this way, RoSA treats layer heterogeneity as a scheduling opportunity rather than only as a pruning signal.

\section{RoSA: Rotational Sparse Adaptation}
\label{sec:method}

RoSA is a memory-efficient fine-tuning framework based on the insight that not all layers need to be updated simultaneously, and that early input-side layers can often be kept fixed during downstream adaptation. Instead of training the full model at every step, RoSA freezes a large input-side prefix, prioritizes one consecutive subset of later layers, and moves this active subset over depth as training progresses. Figure~\ref{fig:rosa_schedule} shows the schedule used in this paper. RoSA has two components. The first is \textit{rotation}: we freeze an early prefix of layers, keep an output-side anchor active, and rotate one trainable block through the remaining depth. The second is \textit{sparse optimization} inside the active block: instead of maintaining optimizer states for all active parameters, we maintain them only for a selected subset. Together, these two components reduce memory at the layer level and at the parameter level. Our gradient analysis and core-only ablation motivate this design by showing that early layers contribute less in our setting, while the middle-to-upper layers carry most of the useful adaptation signal.

In our experiments on Llama-2-7B, we freeze the input embeddings and the first eight Transformer layers, keep the final Transformer layer and output head active, and rotate one consecutive group through the remaining middle-to-upper layers. This is a practical schedule choice rather than a claim that all models must share the same layer profile. The key structural requirement is that the frozen layers form a consecutive prefix, since this is what makes the backward pass shorter.

\textbf{Rotational block training.}
Let a model $\mathcal{M}$ consist of an embedding layer $\mathcal{E}$ and $L$ Transformer layers. We partition the parameters into three sets: a permanently frozen prefix $\mathcal{F}$, an always-active anchor $\mathcal{A}$ near the output, and a rotatable set $\mathcal{R}$ covering the remaining layers. The rotatable set is divided into $K$ consecutive groups, and only one group is active at a time together with the anchor set. Over the full training run, different parts of the network receive updates, but the number of simultaneously trainable layers stays small.
\begin{algorithm}[tb]
\caption{RoSA: Rotational Sparse Adaptation}
\label{alg:rosa}
\begin{algorithmic}[1]
\STATE \textbf{Input:} model parameters $\theta$, training data $\mathcal{D}$, epochs $E$, group size $b$, sparsity density $\rho$, mask refresh interval $\Delta t$
\STATE \textbf{Partition layers:} choose a frozen prefix $\mathcal{F}$, an always-active anchor $\mathcal{A}$, and rotatable layers $\mathcal{R}$
\STATE \textbf{Group rotatable layers:} $\mathcal{R} = G_1 \uplus G_2 \uplus \cdots \uplus G_K$, where each $G_k$ is consecutive and $|G_k|\le b$
\STATE Initialize optimizer states only for parameters that become active

\FOR{$e = 1,\dots,E$}
    \STATE $k \leftarrow 1 + ((e-1) \bmod K)$
    \STATE $\mathcal{S}_e \leftarrow \mathcal{A} \cup G_k$ \COMMENT{active layers for this epoch}
    \STATE Freeze all parameters outside $\mathcal{S}_e$ and discard their optimizer states
    \STATE Initialize fresh optimizer states for newly activated layers \COMMENT{cold start, also when $G_k$ is revisited}
    \STATE Enable gradients only for parameters $\theta_{\mathcal{S}_e}$

    \FOR{$(x,y)\sim\mathcal{D}$}
        \STATE Compute loss $\mathcal{L}(f(x;\theta),y)$ with gradients only through $\theta_{\mathcal{S}_e}$
        \STATE $g_t \leftarrow \nabla_{\theta_{\mathcal{S}_e}}\mathcal{L}(f(x;\theta),y)$
        \IF{$t \bmod \Delta t = 0$}
            \STATE $M_t \leftarrow \textsc{Select}_{\rho}(\theta_{\mathcal{S}_e})$ \COMMENT{NanoAdam mask, e.g., bottom-$\rho$ by $|w|$}
        \ENDIF
        \STATE Update only masked parameters: $\theta_{M_t} \leftarrow \mathrm{AdamW}\!\left(\theta_{M_t}, g_{t,M_t}\right)$
        \STATE Store optimizer states only for $M_t$
    \ENDFOR
\ENDFOR
\end{algorithmic}
\end{algorithm}

In our Llama-2-7B experiments, $\mathcal{F}$ consists of the input embeddings and Layers 0--7, $\mathcal{A}$ consists of the final Transformer layer and output head, and $\mathcal{R}$ consists of Layers 8--30. Unless stated otherwise, we use group size $b=8$, sparsity density $\rho=0.2$, and refresh the sparse mask every $\Delta t=600$ steps. When RoSA is combined with adapters, $\theta_{\mathcal{S}_e}$ denotes the adapter parameters in the active layers; when combined with dense optimization, it denotes the corresponding base-model parameters.

This schedule separates the parameters that are updated \textit{over the full training run} from the parameters that are updated \textit{at any single step}. At each step, RoSA only pays the memory cost of the active block and the output-side anchor. Across epochs, however, the active block moves through depth, so a larger set of layers can receive updates over the course of training. This differs from traditional PEFT methods, which usually keep the same adapter or low-rank subspace trainable throughout training.

\textbf{Optimizer states across rotations.}
When a group leaves the active set, its first- and second-moment estimates are deleted; they are re-initialized from scratch when the group is revisited (a cold start), since retaining them on the GPU would make optimizer memory grow with every visited group. Peak memory thus stays at 34.9\,GiB throughout both cycles on Llama-2-7B, and the training loss decreases across all group transitions without spikes, including the return to the first group in the second cycle (Appendix~\ref{app:stability}).

\textbf{Frozen prefix and truncated backpropagation.}
The frozen input-side prefix is a central part of RoSA. Since the embedding layer and early Transformer layers are never updated, RoSA does not allocate optimizer states or parameter gradients for them. Moreover, gradients do not need to be propagated through this prefix for parameter updates. If the active block begins at layer $l_{\text{start}}$, then all layers below $l_{\text{start}}$ are fixed, so the backward pass can terminate at that boundary. This shortens the effective backward graph, and the saving becomes larger when the active block moves deeper in the model. In our experiments, this effect appears as up to a 27\% reduction in iteration time for later active groups. The same frozen-prefix structure also enables a future system optimization: the prefix output can be precomputed, cached, and compressed, so later training stages can use this representation directly instead of repeatedly recomputing the fixed early layers.
We stress that this optimization is not part of our implementation: all memory and runtime numbers in this paper are obtained with a full forward pass.

\textbf{Sparse optimization within the active block.}
Rotation limits how many layers are active at once, but the active block can still be large enough that optimizer states dominate memory. To reduce this cost further, we apply sparse optimization inside the active block. In our default experiments, we use NanoAdam \citep{zhou2025pay}, which maintains optimizer states only for a selected subset of active parameters.

Let $\theta_{\text{active}}$ denote the parameters in the current active set. Instead of storing first- and second-moment states for all of them, NanoAdam stores states only for a subset $S_t \subset \theta_{\text{active}}$. In our experiments, this subset is chosen by weight magnitude, so that $S_t$ contains the bottom-$k$ parameters by absolute value inside the active block, where $k$ is determined by the sparsity density $\rho$. Only parameters in $S_t$ maintain optimizer states and receive adaptive updates. The remaining active parameters still participate in the forward and backward pass, but they do not carry the full optimizer-state overhead. We refresh this magnitude-based mask periodically because parameter magnitudes change during training. A parameter that is not selected early can later enter the bottom-$k$ set, and a previously selected parameter can leave it. Periodic refreshes therefore prevent the sparse update set from staying fixed for the entire run.

This gives RoSA a second memory-reduction mechanism. Rotation decides which layers are currently trainable, while sparse optimization decides which parameters inside those layers maintain optimizer states. The same rotational schedule could also be combined with other optimizers or parameterizations.

\textbf{Memory and compute perspective.}
RoSA reduces training memory through two multiplicative factors. The first factor is the active block fraction, namely the ratio between the number of active layers and the total model depth. The second factor is the sparsity density inside the active block. If $\Psi$ is the total number of model parameters and $K_{\mathrm{opt}}$ is the number of bytes required per parameter for optimizer states, then standard dense Adam-style fine-tuning scales as $\Psi K_{\mathrm{opt}}$. In contrast, RoSA scales as $(l/L)\Psi \rho K_{\mathrm{opt}}$, where $l$ is the size of the active block, $L$ is the total number of layers, and $\rho$ is the sparsity density inside the active block. Table~\ref{tab:theoretical_memory} summarizes the theoretical optimizer-state scaling of RoSA compared with full fine-tuning and LoRA. Here $d$ denotes a representative hidden dimension and $r$ denotes the LoRA rank. Unless stated otherwise, we use RoSA to refer to our default instantiation with NanoAdam inside the active block.

\begin{table*}[t]
\centering
\setlength{\abovecaptionskip}{2pt}
\setlength{\belowcaptionskip}{2pt}
\vspace{-4pt}

\begin{minipage}[t]{0.31\textwidth}
\vspace{0pt}
\centering
\scriptsize
\setlength{\tabcolsep}{2pt}
\renewcommand{\arraystretch}{0.85}
\captionof{table}{Theoretical optimizer-state memory. RoSA scales with active depth and sparsity.}
\label{tab:theoretical_memory}
\vspace{-4pt}
\resizebox{\linewidth}{!}{
\begin{tabular}{lcc}
\toprule
\textbf{Method} & \textbf{Opt. states} & \textbf{Scale} \\
\midrule
AdamW (FFT) & $\Psi K_{\mathrm{opt}}$ & $O(\Psi)$ \\
LoRA & $\Psi_{\mathrm{adapter}}K_{\mathrm{opt}}$ & $O(dr)$ \\
RoSA & $(l/L)\Psi\rho K_{\mathrm{opt}}$ & $O((\rho l/L)\Psi)$ \\
\bottomrule
\end{tabular}
}
\end{minipage}
\hfill
\begin{minipage}[t]{0.67\textwidth}
\vspace{0pt}
\centering
\scriptsize
\setlength{\tabcolsep}{2.5pt}
\renewcommand{\arraystretch}{0.85}
\captionof{table}{Task accuracy (\%) on Llama-2-7B commonsense reasoning. RoSA stays close to FFT and DoRA while improving over LoRA.}
\label{tab:detailed_results}
\vspace{-4pt}
\resizebox{\linewidth}{!}{
\begin{tabular}{lccccccccc}
\toprule
\textbf{Method} & \textbf{BoolQ} & \textbf{PIQA} & \textbf{SIQA} & \textbf{HellaS} & \textbf{WinoG} & \textbf{ARC-E} & \textbf{ARC-C} & \textbf{OBQA} & \textbf{Avg} \\
\midrule
FFT (AdamW) & 71.8 & 83.1 & 77.5 & 87.5 & 82.0 & 85.2 & 70.6 & 82.2 & 79.99 \\
LoRA & 69.8 & 79.9 & \textbf{79.5} & 83.6 & \textbf{82.6} & 79.8 & 64.7 & 81.0 & 77.61 \\
DoRA & \textbf{71.8} & \textbf{83.7} & 76.0 & 89.1 & \textbf{82.6} & 83.7 & 68.2 & \textbf{82.4} & \textbf{79.70} \\
\midrule
RoSA & 71.0 & 82.9 & 78.7 & \textbf{89.3} & 82.2 & \textbf{83.8} & \textbf{69.9} & 79.2 & 79.61 \\
\bottomrule
\end{tabular}
}
\end{minipage}

\vspace{-10pt}
\end{table*}
This expression captures only the optimizer-state part of the memory saving. RoSA also reduces training memory through its frozen prefix: early layers do not require optimizer states, parameter gradients, or stored activations for weight-gradient computation. The compute benefit comes from the same structure. The full forward pass is still used in our current implementation, but the backward pass does not need to continue through the frozen prefix once the active block is deeper in the network.
As a result, later rotation stages have a shorter backward path and run up to 27\% faster per iteration.
Appendix~\ref{app:memory_breakdown} decomposes the measured peak memory of RoSA on Llama-2-7B into model weights, parameter gradients, optimizer states, and activations.
\section{Experiments}
\label{sec:experiments}

We evaluate RoSA on Llama-2-7B, Qwen-2.5-3B, and Mistral-v0.1-7B across commonsense reasoning and GSM8K. Our main Llama-2-7B evaluation uses eight commonsense reasoning tasks: BoolQ, PIQA, SIQA, HellaSwag, WinoGrande, ARC-Easy, ARC-Challenge, and OpenBookQA.
Unless stated otherwise, we use the default configuration of Section~\ref{sec:method}; additional training details, hyperparameters, and optimizer-specific sparsity ablations are provided in Appendix~\ref{app:experimental_details}.
\begin{wrapfigure}{r}{0.60\textwidth}
\centering
\includegraphics[width=\linewidth]{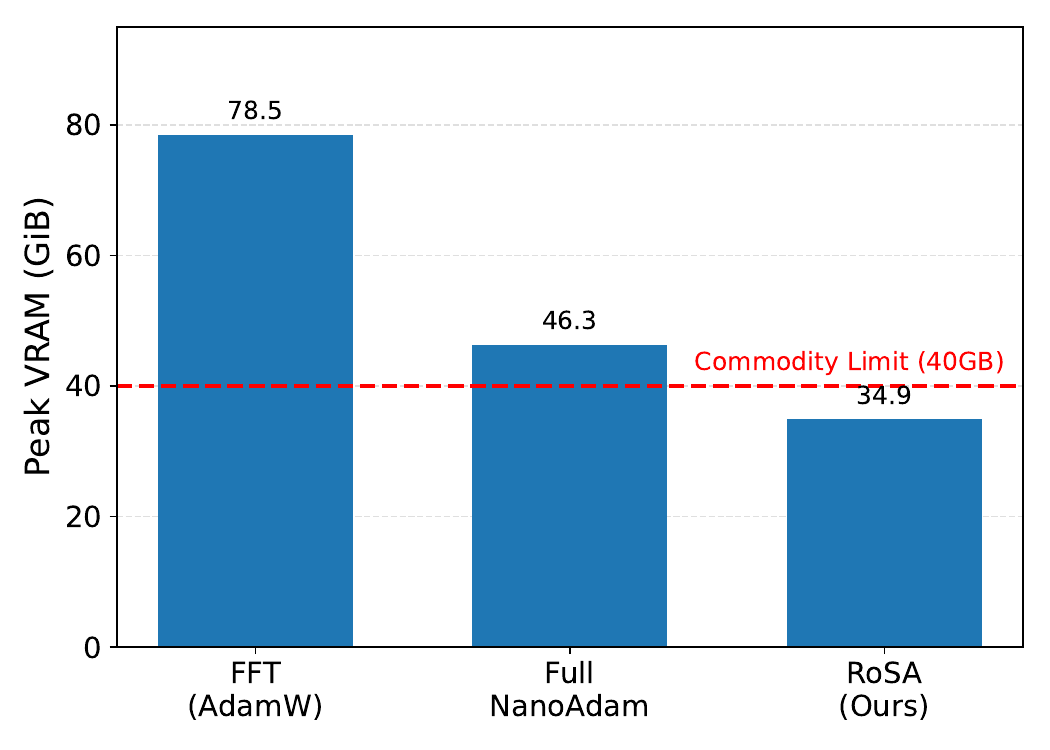}
\caption{Peak VRAM usage. Full fine-tuning with AdamW and full-model sparse optimization remain above the 40\,GB budget, while RoSA fits within it and preserves strong performance.}
\label{fig:vram}
\vspace*{-0.6cm}
\end{wrapfigure}

\noindent\textbf{Llama-2-7B commonsense reasoning.}
Table~\ref{tab:summary_comparison} and Figure~\ref{fig:vram} show the memory trade-off on Llama-2-7B, while Table~\ref{tab:detailed_results} provides the comprehensive per-task accuracy breakdown. Full fine-tuning with AdamW requires 78.5\,GiB, and full-model NanoAdam still requires 46.3\,GiB. RoSA + NanoAdam reduces peak memory to 34.9\,GiB while remaining close to the dense full fine-tuning accuracy. At matched sparsity ($\rho=0.1$ for both), rotation further reduces memory from 46.3\,GiB to 33.9\,GiB, showing that rotation provides an additional reduction beyond sparse optimization alone, and it is this reduction that brings training below the 40\,GB limit.
To assess run-to-run variability, we repeat the default configuration with an independent random seed, which reaches 79.49\% average accuracy compared with 79.61\% in the original run (Appendix~\ref{app:stability}).

\begin{table*}[t]
\centering
\setlength{\abovecaptionskip}{2pt}
\setlength{\belowcaptionskip}{0pt}

\begin{minipage}[t]{0.49\textwidth}
\vspace{0pt}
\centering
\scriptsize
\setlength{\tabcolsep}{2.5pt}
\renewcommand{\arraystretch}{0.9}
\parbox[t][2.2cm][t]{\linewidth}{
\captionof{table}{Peak VRAM, average accuracy, and training time on Llama-2-7B commonsense reasoning. RoSA fits within a 40\,GB budget while staying close to full fine-tuning quality.}
\label{tab:summary_comparison}
}
\vspace{-4pt}
\resizebox{\linewidth}{!}{
\begin{tabular}{lccccc}
\toprule
\textbf{Method} & \textbf{Ep.} & \textbf{Peak VRAM} & \textbf{Fits} & \textbf{Avg Acc.} & \textbf{Time} \\
 &  & \textbf{(GiB)} & \textbf{40\,GB?} &  &  \\
\midrule
FFT (AdamW) & 3 & 78.5 & No & 79.99 & 4.37h \\
Full NanoAdam ($\rho=0.1$) & -- & 46.3 & No & 81.71 & -- \\
RoSA, 1 cycle & 3 & 34.9 & Yes & 79.10 & \textbf{3.5h} \\
RoSA, 2 cycles & 6 & \textbf{34.9} & \textbf{Yes} & 79.61 & 6.5h \\
\bottomrule
\end{tabular}
}
\renewcommand{\arraystretch}{1.0}
\end{minipage}
\hfill
\begin{minipage}[t]{0.49\textwidth}
\vspace{0pt}
\centering
\scriptsize
\setlength{\tabcolsep}{3pt}
\renewcommand{\arraystretch}{0.9}
\parbox[t][2.2cm][t]{\linewidth}{

\captionof{table}{Peak VRAM and GSM8K accuracy under a tight GPU budget. RoSA-LoRA remains close to plain LoRA and outperforms BAdam while using less memory.}
\label{tab:budget_results}
}
\vspace{-4pt}
\resizebox{\linewidth}{!}{
\begin{tabular}{lccc}
\toprule
\textbf{Method} & \textbf{Active Params} & \textbf{Peak VRAM (GB)} & \textbf{GSM8K Acc.} \\
\midrule
LoRA + AdamW & $\sim$34M & 16.20 & 36.69 \\
BAdam & $\sim$202M & 16.7 & 24.48 \\
RoSA-LoRA & $\sim$37M & \textbf{14.6} & 35.25 \\
\bottomrule
\end{tabular}
}
\renewcommand{\arraystretch}{1.0}
\end{minipage}

\vspace{-8pt}
\end{table*}
\noindent\textbf{Generalization across models and tasks.}
The original Llama-2-7B commonsense setting is only one test case. Table~\ref{tab:generalization_results} extends the commonsense evaluation beyond Llama-2-7B. RoSA transfers to Qwen-2.5-3B and Mistral-v0.1-7B, remaining close in accuracy while using substantially less memory, reducing peak VRAM from 22.48\,GB to 14.37\,GB on Qwen-2.5-3B and from 45.30\,GB to 33.28\,GB on Mistral-v0.1-7B.

\begin{table*}[t]
\centering
\begin{minipage}[t]{0.49\textwidth}
\centering
\scriptsize
\parbox[t][1.9cm][t]{\linewidth}{
\captionof{table}{Accuracy and peak VRAM across models on commonsense reasoning. RoSA transfers beyond the original Llama-2-7B setting and remains close to full-model NanoAdam baselines.}
\label{tab:generalization_results}
}
\resizebox{\linewidth}{!}{
\begin{tabular}{llcc}
\toprule
\textbf{Model} & \textbf{Method} & \textbf{Acc. (\%)} & \textbf{Peak VRAM (GB)} \\
\midrule
Qwen-2.5-3B & NanoAdam & 83.47 & 22.48 \\
Qwen-2.5-3B & RoSA + NanoAdam & 82.91 & \textbf{14.37} \\
\midrule
Mistral-v0.1-7B & NanoAdam & 84.25 & 45.30 \\
Mistral-v0.1-7B & RoSA + NanoAdam & 81.86 & \textbf{33.28} \\
\bottomrule
\end{tabular}
}
\end{minipage}
\hfill
\begin{minipage}[t]{0.49\textwidth}
\vspace{0pt}
\centering
\small
\parbox[t][1.9cm][t]{\linewidth}{
\captionof{table}{Average accuracy (\%) with Llama-2-7B on 8 commonsense reasoning tasks and wall-clock time for different active group sizes. Group size 8 gives the best balance in our experiments.}
\label{tab:group_size_ablation}
}
\resizebox{\linewidth}{!}{
\begin{tabular}{lccc}
\toprule
\textbf{Active Group Size} & \textbf{Epochs} & \textbf{Accuracy (\%)} & \textbf{Time (h)} \\
\midrule
4  & 12 & 77.92 & 22.46 \\
8  & 6  & \textbf{79.61} & \textbf{6.50} \\
12 & 4  & 79.33 & 7.22 \\
\bottomrule
\end{tabular}
}
\end{minipage}
\end{table*}
\noindent\textbf{Composability with PEFT under tight memory budgets.}
RoSA can also be combined with adapter-style methods rather than replacing them. On Qwen-2.5-3B commonsense reasoning, plain DoRA reaches 81.58\% average accuracy with 16.42\,GB peak VRAM, while RoSA-DoRA reaches 81.79\% and reduces peak VRAM to 12.87\,GB in the first epoch, with later rotation stages using 11.04\,GB and 9.44\,GB; the results are shown in Table~\ref{tab:rosa_dora_qwen}.
Over three random seeds, RoSA-DoRA reaches $81.83 \pm 0.32\%$ (mean and standard deviation), on par with or slightly above the 81.58\% of plain DoRA. Since adapter optimizer states are already small, the memory saving of RoSA-DoRA comes mostly from removing adapter gradients and states in inactive layers and from not storing activations for weight gradients below the active block, which is also why memory keeps decreasing as the active group moves deeper.
We observe a similar memory-saving pattern in the low-memory LoRA setting on GSM8K. Under a similar active-adapter budget, plain LoRA with AdamW updates about 34M parameters and reaches 36.69\% at 16.20\,GB peak VRAM, while RoSA-LoRA updates about 37M parameters and reaches 35.25\% at 14.6\,GB peak VRAM.
Table~\ref{tab:budget_results} also compares against BAdam, which reaches only 24.48\% at about 16.7\,GB under the same consumer-GPU budget. Together, the RoSA-DoRA and RoSA-LoRA results show that RoSA is complementary to PEFT parameterizations: it can reduce memory on top of different adapter-style methods while preserving most of their adaptation performance.
Since neither of these hybrids uses NanoAdam, they also show that rotation reduces memory on its own, without sparse optimization inside the active block.

\begin{table*}[t]
\centering
\setlength{\abovecaptionskip}{2pt}
\setlength{\belowcaptionskip}{0pt}
\begin{minipage}[t]{0.56\textwidth}
\vspace{0pt}
\centering
\scriptsize
\setlength{\tabcolsep}{2.5pt}
\renewcommand{\arraystretch}{0.9}
\captionof{table}{Layer-wise and memory-efficient baselines on Llama-2-7B commonsense reasoning. $^\dagger$Does not fit into 40\,GB GPUs at cutoff length 256; run on 80\,GB GPUs.}
\label{tab:baselines}
\vspace{2pt}
\resizebox{\linewidth}{!}{
\begin{tabular}{lccccc}
\toprule
\textbf{Method} & \textbf{Cutoff} & \textbf{Ep.} & \textbf{Peak VRAM} & \textbf{Avg Acc.} & \textbf{Time} \\
 & \textbf{length} &  & \textbf{(GiB)} &  &  \\
\midrule
FFT (AdamW) & 256 & 3 & 78.5 & 79.99 & 4.37h \\
GaLore (8-bit, $r=64$)$^\dagger$ & 64 & 3 & 41.44 & 31.28 & 14.05h \\
LISA (8 active layers) & 256 & 3 & 30.71 & 70.05 & 3.8h \\
\midrule
RoSA, group size 6 & 256 & -- & 30.93 & 78.50 & -- \\
RoSA, group size 8, 1 cycle & 256 & 3 & 34.9 & 79.10 & \textbf{3.5h} \\
RoSA, group size 8, 2 cycles & 256 & 6 & 34.9 & \textbf{79.61} & 6.5h \\
\bottomrule
\end{tabular}
}
\renewcommand{\arraystretch}{1.0}
\end{minipage}
\hfill
\begin{minipage}[t]{0.41\textwidth}
\vspace{0pt}
\centering
\scriptsize
\setlength{\tabcolsep}{2.5pt}
\renewcommand{\arraystretch}{0.9}
\captionof{table}{Static layer subsets versus rotation on Llama-2-7B (anchor active, NanoAdam with $\rho=0.2$, 3 epochs).}
\label{tab:static_subsets}
\vspace{2pt}
\resizebox{\linewidth}{!}{
\begin{tabular}{lcc}
\toprule
\textbf{Trainable layers} & \textbf{Peak VRAM (GiB)} & \textbf{Avg Acc.} \\
\midrule
Static, Layers 8--15 & 26.2 & 78.32 \\
Static, Layers 16--23 & 25.9 & 30.07 \\
Static, Layers 24--30 & 25.5 & 33.68 \\
Static, Layers 8--23 & 31.5 & 78.07 \\
\midrule
RoSA, 1 cycle (rotating 8--30) & 34.9 & \textbf{79.10} \\
\bottomrule
\end{tabular}
}
\renewcommand{\arraystretch}{1.0}
\end{minipage}
\vspace{-8pt}
\end{table*}

\noindent\textbf{Comparison with layer-wise and memory-efficient baselines.}
We further compare RoSA with LISA \citep{pan2024lisa} and GaLore \citep{zhao2024galore} on Llama-2-7B (Table~\ref{tab:baselines}; details in Appendix~\ref{app:experimental_details}). With the best of five learning rates, LISA reaches 70.05\% at 30.71\,GiB, while RoSA reaches 79.10\% with the same number of epochs and 79.61\% after two cycles. At matched memory, RoSA with group size 6 uses 30.93\,GiB and still reaches 78.50\%, 8.45 points above LISA.
GaLore does not fit into 40\,GB GPUs at cutoff length 256 in our data-parallel setup, since dense gradients are still computed for all layers before the low-rank projection; with rank 64 and cutoff length 64 it peaks at 41.44\,GiB and reaches 31.28\% after 14.05 hours. Since the shortened context likely explains much of this low accuracy, we read this comparison mainly in terms of memory.

\noindent\textbf{Schedule sensitivity and training dynamics.}
We evaluate whether RoSA depends on one narrow schedule choice. A single-cycle schedule finishes in 3 epochs and reaches 79.10\% average accuracy in about 3.5 hours, which is shorter than the 4.37 hours of full fine-tuning with AdamW, while the default double-cycle schedule reaches 79.61\% in about 6.5 hours. Table~\ref{tab:summary_comparison} shows that the second cycle improves accuracy, but the convergence behavior suggests that RoSA does not always need to be trained for the full double-cycle budget: the evaluation loss decreases rapidly in the first cycle and only slightly afterwards. For example, the minimum evaluation loss improves from 0.7484 after one cycle to 0.7349 after two cycles, indicating that most of the convergence already happens in the shorter run. Importantly, RoSA also becomes cheaper within a rotation cycle: later active groups are up to 27\% faster per iteration because the backward path is shorter. Thus, RoSA exposes a practical trade-off: one can stop after a shorter schedule for faster training, or continue rotating for modest additional accuracy while staying under the 40\,GB memory budget. Table~\ref{tab:group_size_ablation} further shows that group size 8 gives the best balance in our experiments, while nearby settings remain competitive.

\begin{figure*}[t]
    \centering
    \setlength{\abovecaptionskip}{2pt}
    \setlength{\belowcaptionskip}{0pt}
    \includegraphics[width=\textwidth]{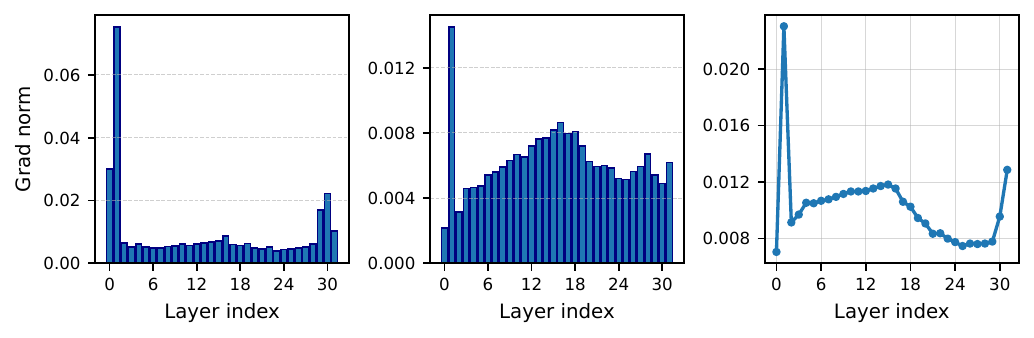}
    \caption{Layer-wise gradient norm analysis of Llama-2-7B during full fine-tuning. Left: Step 100, where activity is dominated by the input and head. Middle: Step 1100, where a ``reasoning hump'' emerges in Layers 8--18 while Layers 2--7 remain flat. Right: the average gradient profile over the first epoch, which highlights inefficient updates in deeper Layers 20--29. RoSA exploits this structure by freezing stable early Layers 0--7 and rotating updates through the more active middle layers.}
    \label{fig:gradient_analysis}
    \vspace{-10pt}
\end{figure*}

\noindent\textbf{Layer-wise gradient dynamics.}
To better understand why a rotational schedule works, we track layer-wise gradient norms during dense fine-tuning on the Llama-2-7B commonsense setting. Figure~\ref{fig:gradient_analysis} shows three views of this signal: the layer-wise gradient norms at an early training step, the layer-wise gradient norms at a later training step, and the per-layer gradient norm averaged over all steps in the first epoch.

The main pattern is that adaptation pressure is not uniform over depth. The earliest layers are comparatively quiet, while the middle-to-upper layers carry most of the training signal. This supports the main design choice behind RoSA: freezing an early prefix and spending the available memory budget on a rotating block in the more active part of the network. We do not claim that the exact same gradient profile must appear in every model or every task. Instead, we use this analysis as a practical guide for schedule design.
Short gradient probes on Mistral-v0.1-7B and Qwen-2.5-3B show the same coarse structure, with a comparatively flat region up to about Layer 11 and higher gradient norms in the middle-to-upper layers (Appendix~\ref{app:grad_probes}).

A second pattern is that the final layer remains consistently important. This motivates keeping an output-side anchor active throughout training. The anchor helps maintain a stable interface to the output head while the trainable block rotates through the internal layers. In our experiments, this simple structure works well across different schedules and model families.

\noindent\textbf{Cumulative capacity versus instantaneous cost.}
A useful way to understand RoSA is to separate what happens at one training step from what happens over the full training run. At any single step, only the active block and the anchor carry trainable parameters and optimizer states. This keeps the instantaneous memory cost low. In our main Llama-2-7B setting, optimizer states are maintained for about 7.4\% of the model parameters at one time (about 495.5M parameters; see Appendix~\ref{app:memory_breakdown}). This is the quantity that determines the instantaneous optimizer-state footprint.

Over time, however, the active block moves through the network, so different parts of the model receive updates at different stages. To measure this cumulative adaptation capacity, we track the union of parameters that enter the sparse update set over the full training run. Under this definition, RoSA updates about 18.8\% of the model parameters across training, corresponding to roughly 1.26 billion parameters on Llama-2-7B.
For comparison, typical LoRA configurations train well below 1\% of the model parameters over the full run.

This is the core trade-off of the method. RoSA keeps the instantaneous footprint small, but it is not restricted to the same trainable subspace at every step. Instead, it reuses a fixed memory budget over different parts of the model across time. This allows RoSA to adapt a larger cumulative set of parameters over time than PEFT methods, while maintaining a memory footprint significantly lower than dense fine-tuning.

This view also helps explain why the second rotation cycle improves results. After later groups adapt, returning to earlier groups gives them a chance to realign to the updated feature space. This is consistent with the gain we observe when moving from the shorter single-cycle schedule to the longer double-cycle schedule.

\noindent\textbf{Rotation versus static layer subsets.}
Could a single, carefully chosen static subset of layers match rotation? We train static blocks with the same output anchor, optimizer, and 3-epoch budget as single-cycle RoSA (Table~\ref{tab:static_subsets}). The middle block (Layers 8--15) is the strongest static choice with 78.32\%, while training only deeper blocks fails in our setting. Doubling the block to Layers 8--23 does not help (78.07\%), whereas single-cycle RoSA, which visits each block once, reaches 79.10\%. The gain of RoSA thus comes from moving the active block over depth rather than from the capacity of any single block.

\noindent\textbf{Targeting the reasoning core.}
The gradient analysis suggests that the most useful adaptation signal is concentrated in the middle part of the network. To test how much of the full RoSA schedule is actually needed, we run a core-only ablation that rotates only through Layers 8--23 while keeping the early prefix, the deeper Layers 24--30, and the output-side anchor fixed.

Table~\ref{tab:core_ablation} shows that this narrower schedule performs almost identically to full RoSA: average accuracy changes from 79.61\% to 79.56\%. This supports the view that RoSA is a way to allocate memory over depth, and that most of the useful adaptation signal in this setting is concentrated in the middle layers once the output-side anchor remains active.
\begin{table}[t]
\centering
\setlength{\abovecaptionskip}{5pt}
\setlength{\belowcaptionskip}{8pt}
\footnotesize
\setlength{\tabcolsep}{8pt}
\renewcommand{\arraystretch}{0.86}
\caption{RoSA-DoRA on Qwen-2.5-3B commonsense reasoning. RoSA-DoRA maintains similar accuracy while reducing VRAM across rotation stages. The last row reports the mean and standard deviation over three seeds.}
\label{tab:rosa_dora_qwen}
\vspace{-5pt}
\begin{tabular}{lccccc}
\toprule
\textbf{Method} & \textbf{Ep.} & \textbf{Acc.} & \textbf{Peak} & \textbf{Later} & \textbf{Runtime} \\
 &  & \textbf{(\%)} & \textbf{VRAM} & \textbf{VRAM} &  \\
\midrule
DoRA & 3 & 81.58 & 16.42 & -- & 15h 28m \\
RoSA-DoRA & 6 & \textbf{81.79} & \textbf{12.87} & 11.04/9.44 & 23h 40m \\
RoSA-DoRA (3 seeds) & 6 & 81.83$\pm$0.32 & 12.87 & -- & -- \\
\bottomrule
\end{tabular}
\renewcommand{\arraystretch}{1.0}
\end{table}

\begin{table}[t]
\centering
\scriptsize
\renewcommand{\arraystretch}{0.82}
\caption{Core-only RoSA ablation. Restricting rotation to Layers 8--23 gives results almost identical to the full RoSA schedule. The average is computed over the full 8-task suite.}
\label{tab:core_ablation}
\vspace{-4pt}
\begin{tabular*}{0.78\linewidth}{@{\extracolsep{\fill}}lccc}
\toprule
\textbf{Task} & \textbf{Full} & \textbf{Core} & \textbf{$\Delta$} \\
\midrule
OpenBookQA & 79.2 & 79.4 & +0.2 \\
PIQA & 82.9 & 83.1 & +0.2 \\
WinoGrande & 82.2 & 82.2 & 0.0 \\
ARC-Easy & 83.8 & 83.6 & -0.2 \\
HellaSwag & 89.3 & 89.0 & -0.3 \\
BoolQ & 71.0 & 70.7 & -0.3 \\
\midrule
Average & 79.61 & 79.56 & -0.05 \\
\bottomrule
\end{tabular*}
\vspace{-8pt}
\renewcommand{\arraystretch}{1.0}
\end{table}

Appendix~\ref{app:sparsity_frequency} includes an additional sparsity-frequency ablation showing that sparsity more aggressive than our default requires more frequent mask refreshes to avoid performance loss.

\section*{Limitations}

RoSA reduces memory and shortens the backward pass.
It also has the potential to realize a shorter forward pass by defining the activations of the last frozen layer as new data samples.
We do not implement this in the present work, so all memory and runtime numbers we report are obtained with a full forward pass.
However, such caching can increase the memory demands of the training data and slow down data loading.
For Llama-2-7B with cutoff length 256, the bf16 hidden states at the prefix boundary take up to 2\,MiB per sequence (about 200\,GiB per $10^5$ sequences), and offloading them uncompressed to CPU memory would be limited by transfer bandwidth.
To overcome this issue, the data would need to be compressed, which adds computation.
A practical implementation would likely need fast INT8/FP8 activation compression integrated at the library level (C++/CUDA).
This would need to be achieved dynamically at later training rounds when new layers are frozen permanently.

Furthermore, RoSA introduces additional scheduling choices such as active-group size, number of rotation cycles, and sparsity-refresh interval. Our experiments show that RoSA is relatively robust to such choices, but they still affect the time--accuracy trade-off and may need retuning for new models or tasks.
In particular, we use the same frozen prefix (Layers 0--7) for all models without ablating its length, although the probes in Appendix~\ref{app:grad_probes} suggest that longer prefixes may be possible for some models. Warm-starting revisited groups from offloaded optimizer states is another option we have not explored.

While RoSA reduces optimizer-state memory and shortens the effective backward path in later training stages, it can still require more training epochs to converge and therefore take longer in wall-clock time than unconstrained dense fine-tuning when the latter fits in memory. RoSA is therefore most useful when memory is the main bottleneck, rather than when raw training speed is the top priority.

Our empirical study is broader than a single-model, single-task evaluation, but is still limited to models up to 7B parameters and a small set of tasks; we do not evaluate instruction tuning or multimodal fine-tuning, and we report repeated runs only for the main configurations. The layer-activity patterns that motivate the schedule should be understood as a practical design guide rather than a universal law. We also do not combine RoSA with activation checkpointing (none of our runs use it), ZeRO-style offloading, or quantized backbones as in QLoRA, which act on other parts of the memory footprint and could be stacked with RoSA.

Finally, we focus on memory-efficient adaptation rather than deployment behavior. This paper does not study downstream fairness, safety, or misuse risks beyond the standard model and dataset choices used in our experiments.

\section{Conclusions and Discussion}
\label{sec:conclusion}

We introduced RoSA, a memory-aware fine-tuning framework that freezes a large input-side prefix and rotates trainable layer groups over the remaining depth. The key idea is to reuse a fixed memory budget over different parts of the model across time, while avoiding optimizer states and parameter gradients for the frozen prefix. This decouples the set of parameters that can be adapted over the full run from the set that must be active at any single step.

Across our experiments, RoSA stays close to dense full fine-tuning while operating under a much smaller memory budget. On Llama-2-7B, it reduces peak VRAM from 78.5\,GiB for full fine-tuning with AdamW to 34.9\,GiB while reaching 79.61\% average accuracy on the commonsense reasoning suite. We also show that the method transfers beyond the original setting to Qwen-2.5-3B, Mistral-v0.1-7B, and GSM8K, remains stable across reasonable schedule choices, and composes with PEFT methods such as DoRA and LoRA.
Rotating the active block also outperforms training static layer subsets, and RoSA compares favorably with LISA, BAdam, and GaLore while using comparable or less memory.

The main advantage of RoSA is that it exploits the depth structure of Transformer models rather than only changing the optimizer or the trainable parameterization: sparse optimization reduces optimizer-state cost inside the active block, rotation reduces the number of layers active at any one time, and the frozen prefix shortens the backward path and could in future enable cached or compressed prefix activations. This makes RoSA a modular layer-scheduling principle that can be combined with dense, sparse, or adapter-style fine-tuning. Instead of treating the hardware budget as a fixed constraint on a fixed trainable subspace, RoSA allocates it across depth over time, which is especially useful in shared compute environments, where later rotation stages require less memory and could run on lower-memory devices. We believe this makes rotational adaptation a promising direction for efficient LLM fine-tuning that is complementary to traditional parameter-efficient fine-tuning.

\begin{ack}
The authors gratefully acknowledge the Gauss Centre for Supercomputing e.V. for funding this project by providing computing time on the GCS Supercomputer JUWELS at J\"ulich Supercomputing Centre (JSC). We also gratefully acknowledge funding from the European Research Council (ERC) under the Horizon Europe Framework Programme (HORIZON) for proposal number 101116395 SPARSE-ML. Computing resources were also provided by the CISPA Helmholtz Center for Information Security. The authors declare no competing interests.
\end{ack}

\bibliographystyle{plain}
\bibliography{references}

\appendix
\newpage

\section{Additional Experimental Details}
\label{app:experimental_details}

\textbf{Models and tasks.}
We evaluate RoSA on three open LLM families: Llama-2-7B \citep{touvron2023llama}, Mistral-v0.1-7B, and Qwen-2.5-3B. Our main evaluation uses eight tasks from the commonsense reasoning benchmark: BoolQ, PIQA, SIQA, HellaSwag, WinoGrande, ARC-Easy, ARC-Challenge, and OpenBookQA. We report average accuracy over these eight tasks. To test whether the method also transfers beyond multiple-choice reasoning, we additionally evaluate on GSM8K, which requires generative mathematical reasoning. This setup tests RoSA across model families, model scales, and task types.

\textbf{Baselines.}
We compare RoSA against both dense and parameter-efficient baselines. On the Llama-2-7B commonsense suite, we compare against full fine-tuning with AdamW, LoRA \citep{hu2021lora}, and DoRA \citep{liu2024dora}. When dense AdamW is not feasible within the target memory budget, we use full-model NanoAdam as the strongest dense baseline designed for reduced memory usage. To study the low-memory adapter regime, we also evaluate plain LoRA and a RoSA-LoRA hybrid on GSM8K under closely matched active-adapter budgets. For a direct block-wise dense baseline under a consumer-GPU memory regime, we additionally compare against BAdam \citep{chen2024badam} on GSM8K.
On the Llama-2-7B commonsense suite, we further compare against LISA \citep{pan2024lisa}, GaLore \citep{zhao2024galore}, and static layer subsets. These baselines use global batch size 128, micro-batch size 1, and 3 epochs. For LISA, we use 8 active layers and select the learning rate from a five-point sweep, in which $1\times10^{-4}$ performed best. For GaLore, we use the 8-bit variant with rank 64 and learning rate $1\times10^{-4}$; since it does not fit into 40\,GB GPUs at a cutoff length of 256, we reduce the cutoff length to 64 and run it on A100 80GB GPUs. The static-subset baselines keep the output anchor active and use NanoAdam with $\rho=0.2$, exactly as RoSA, but never move the trainable block.

\textbf{RoSA configuration.}
Unless stated otherwise, RoSA freezes the input embeddings and Layers 0--7, keeps the final Transformer layer and the output head always active, and rotates one consecutive trainable group through Layers 8--30. Our default active-group size is 8 layers. We use a cyclic 6-epoch schedule so that each rotating group is visited twice. Inside the active block, we use NanoAdam with learning rate $1\times 10^{-4}$, $\beta_1=0.9$, $\beta_2=0.95$, and sparsity density $\rho=0.2$. The sparse mask is refreshed every 600 steps unless noted otherwise.

For Mistral-v0.1-7B and Qwen-2.5-3B, we keep the same frozen prefix (Layers 0--7) and the same type of output anchor (final Transformer layer and head), and use active-group sizes of 8 and 9, respectively. Optimizer states of a rotating group are deleted when it becomes inactive and re-initialized when it is revisited (Section~\ref{sec:method}).

\textbf{Ablations.}
We study the effect of the active-group size, the rotation schedule length, and the sparsity-update frequency. For the group-size ablation, we evaluate active groups of size 4, 8, and 12. Since changing the group size changes how many groups cover the model, we adjust the total number of epochs to keep rotational coverage comparable: 12 epochs for group size 4, 6 epochs for group size 8, and 4 epochs for group size 12. We also compare a single-cycle schedule of 3 epochs against the default double-cycle schedule of 6 epochs. In a separate sparsity ablation, we reduce the density to $\rho=0.1$ and vary the mask refresh interval to study the trade-off between lower optimizer-state memory and adaptation quality.

\textbf{Implementation details and reporting.}
We implement RoSA in PyTorch using the Hugging Face \texttt{transformers} library. Unless noted otherwise, experiments are run on a cluster with 4$\times$NVIDIA A100 40GB GPUs. For the Llama-2-7B commonsense experiments used in the time and memory analysis, we use micro-batch size 1, total batch size 128, and no gradient checkpointing. We report task accuracy, average accuracy across the commonsense suite, peak VRAM, and wall-clock training time where relevant. For comparisons in the low-memory regime, we also report the number of simultaneously active parameters when this helps clarify how different methods use a fixed memory budget.
Baselines whose peak memory exceeds 40\,GB (full fine-tuning with AdamW, full-model NanoAdam, and GaLore) are run on A100 80GB GPUs. None of the RoSA runs use activation checkpointing.

\section{Interplay of Sparsity and Update Frequency}
\label{app:sparsity_frequency}

RoSA combines layer rotation with sparse optimization inside the active block, so it is important to understand how these two parts interact. To study this, we reduce the NanoAdam density from $\rho=0.2$ to $\rho=0.1$ and compare two mask refresh intervals: a slower update schedule with $\Delta t=600$ steps and a faster update schedule with $\Delta t=100$ steps.

Table~\ref{tab:sparsity_ablation} shows a clear pattern. When the density is reduced to $\rho=0.1$ and the mask is refreshed only every 600 steps, performance drops noticeably, with the largest degradation appearing on HellaSwag. This suggests that under extreme sparsity, the active update set becomes stale before it can track the changing training signal. When the same sparse budget is combined with a faster refresh interval of 100 steps, most of this loss is recovered.

This result shows that spatial sparsity and update frequency are closely linked. Lower density can still work, but only when the sparse mask is allowed to adapt more quickly. In other words, very sparse optimization can remain competitive, but it needs a faster mechanism for following the useful part of the gradient landscape. In practice, this comes with extra overhead from more frequent mask recomputation. For this reason, we use $\rho=0.2$ with $\Delta t=600$ as the default setting, since it gives the best overall balance in our experiments.

\begin{table}[t]
\centering
\small
\caption{Task accuracy (\%) under different sparsity-update settings. Lower density hurts performance when the sparse mask is refreshed slowly, but much of this loss can be recovered by refreshing the mask more often.}
\label{tab:sparsity_ablation}
\resizebox{\linewidth}{!}{
\begin{tabular}{lccc}
\toprule
\textbf{Task} &
\shortstack{\textbf{Baseline}\\$\rho=0.2,\ \Delta t=600$} &
\shortstack{\textbf{Low Freq.}\\$\rho=0.1,\ \Delta t=600$} &
\shortstack{\textbf{High Freq.}\\$\rho=0.1,\ \Delta t=100$} \\
\midrule
ARC-C   & \textbf{69.9} & 68.3 & 67.5 \\
ARC-E   & \textbf{83.8} & 82.8 & 83.2 \\
BoolQ   & \textbf{71.0} & 69.7 & 68.8 \\
HellaS  & \textbf{89.3} & 74.9 & 87.9 \\
OBQA    & \textbf{79.2} & 77.8 & 78.0 \\
PIQA    & \textbf{82.9} & 82.5 & \textbf{82.9} \\
SIQA    & 78.7 & 78.1 & \textbf{78.8} \\
WinoGr  & \textbf{82.2} & 79.8 & 80.7 \\
\midrule
\textbf{Average} & \textbf{79.6} & 76.7 & 78.5 \\
\bottomrule
\end{tabular}
}
\end{table}

\section{Peak Memory Breakdown}
\label{app:memory_breakdown}

To separate the contributions of rotation and sparse optimization, Table~\ref{tab:memory_breakdown} decomposes the peak memory of RoSA + NanoAdam ($\rho=0.2$, group size 8) on Llama-2-7B with micro-batch size 1 and cutoff length 256. At a single step, NanoAdam tracks about 495.5M parameters, or 7.35\% of the 6.74B model parameters. These comprise the sparse subsets of the active Transformer layers, the dense output head, and the normalization parameters. Parameter gradients are only allocated for the active group, the anchor layer, and the head. Optimizer states take only 2.52\,GiB, since NanoAdam stores the moment estimates in bfloat16 together with 16-bit indices. Most of the remaining memory is taken by the frozen model weights and by activations.

These numbers also clarify how the two mechanisms of RoSA interact. At matched sparsity ($\rho=0.1$), sparse optimization alone reduces the peak memory from 78.5\,GiB for full fine-tuning with AdamW to 46.3\,GiB for full-model NanoAdam, and rotation removes a further 12.4\,GiB, down to 33.9\,GiB. Rotation is therefore the component that brings training below the 40\,GB limit in this setting.

\begin{table}[h]
\centering
\small
\caption{Breakdown of the peak memory of RoSA + NanoAdam on Llama-2-7B commonsense reasoning ($\rho=0.2$, group size 8, micro-batch size 1, cutoff length 256).}
\label{tab:memory_breakdown}
\begin{tabular}{lc}
\toprule
\textbf{Component} & \textbf{Memory (GiB)} \\
\midrule
Model weights (bf16, all layers) & 12.55 \\
Parameter gradients (active group, anchor layer, head) & 3.64 \\
Optimizer states (NanoAdam, $\sim$495.5M tracked parameters) & 2.52 \\
Activations and other allocations & 13.91 \\
\midrule
Peak allocated tensor memory (PyTorch) & 32.62 \\
CUDA context and caching allocator & $\sim$2.3 \\
\midrule
Peak VRAM (system) & 34.9 \\
\bottomrule
\end{tabular}
\end{table}

\section{Layer-wise Gradient Profiles on Other Models}
\label{app:grad_probes}

\begin{figure}[h]
\centering
\includegraphics[width=0.8\textwidth]{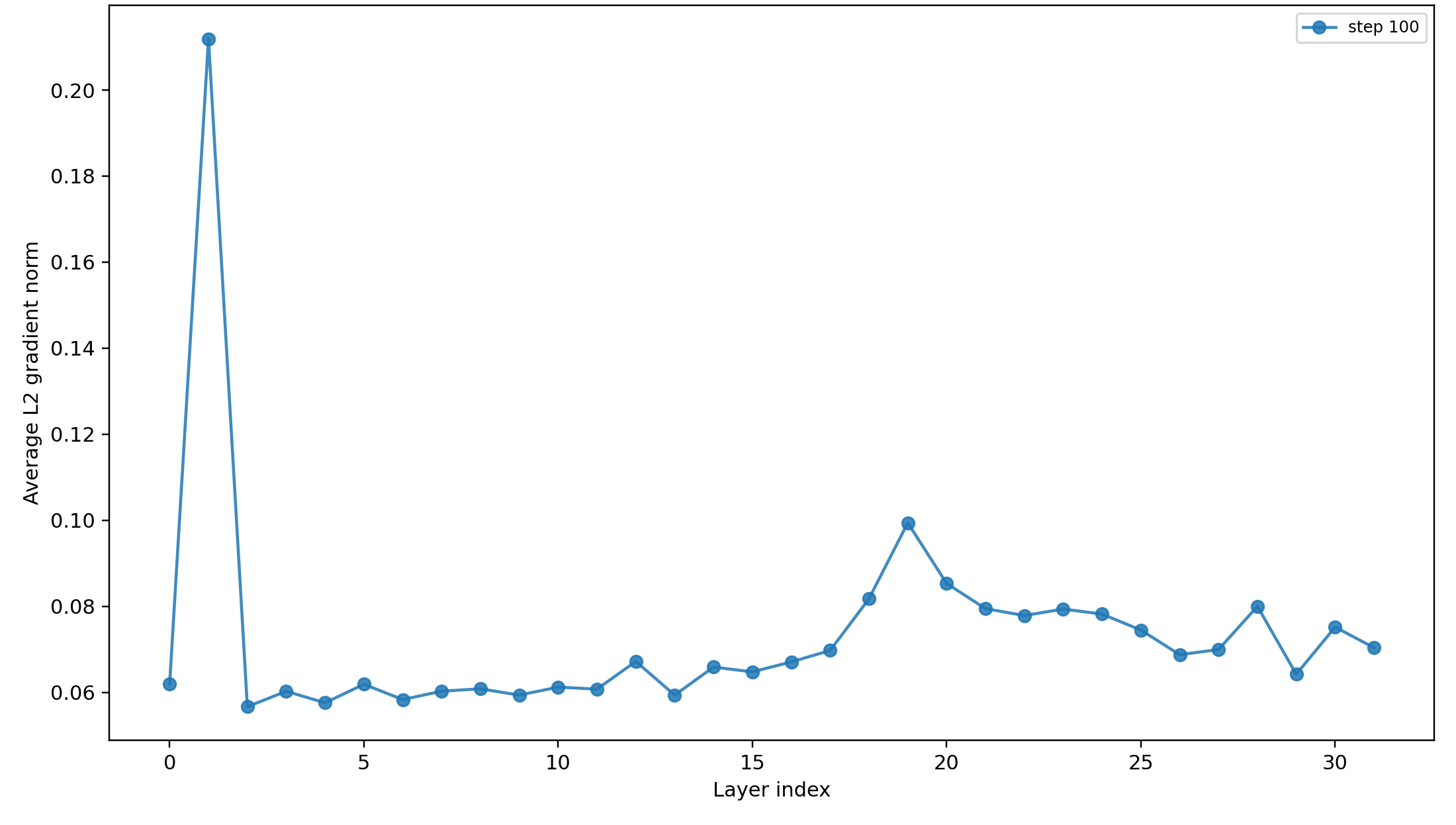}\\
{\small (a) Mistral-v0.1-7B, step 100}\\[6pt]
\includegraphics[width=0.8\textwidth]{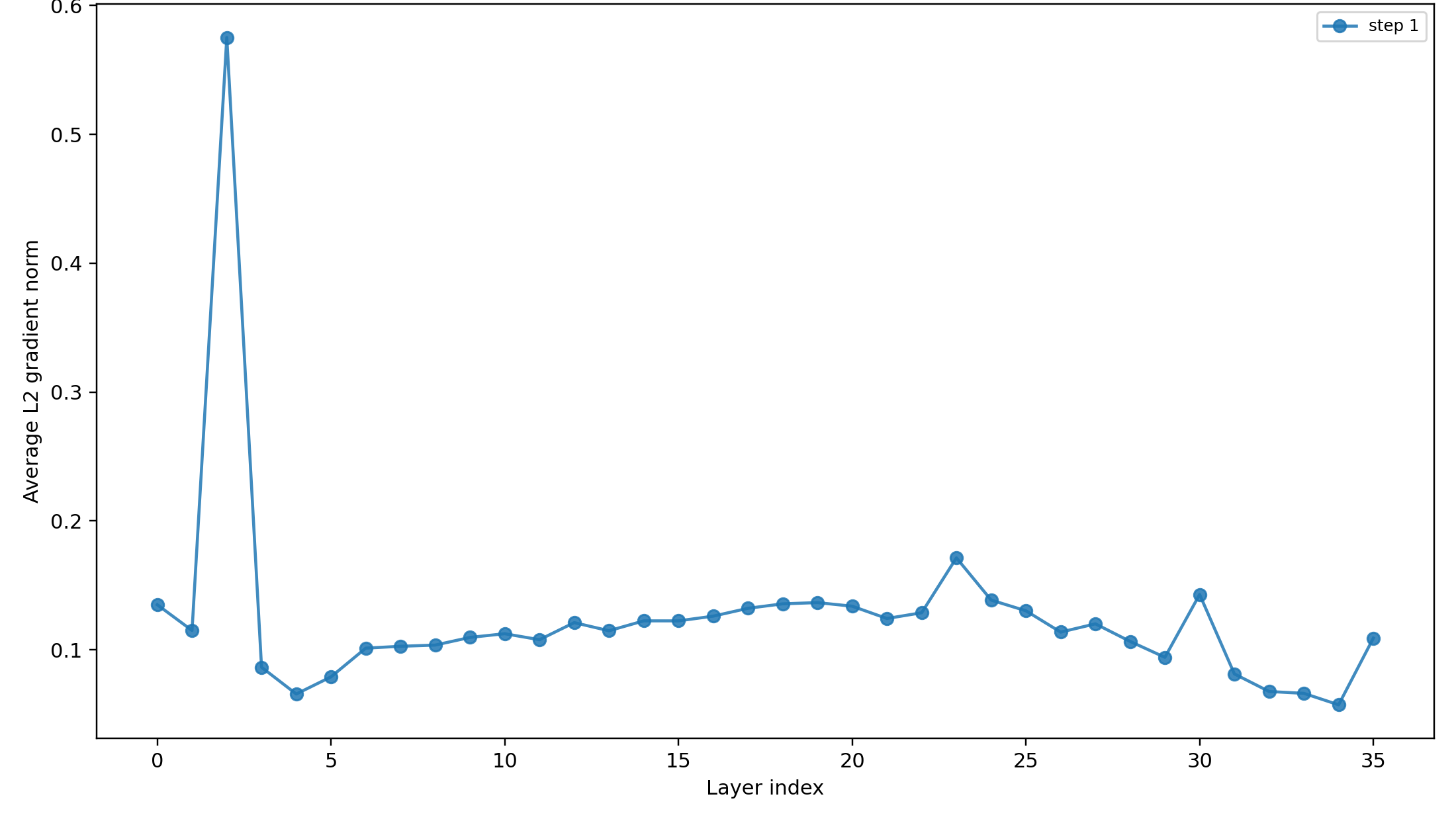}\\
{\small (b) Qwen-2.5-3B, step 1}
\caption{Average per-layer $\ell_2$ gradient norms from short dense fine-tuning probes. Axis scales differ between panels.}
\label{fig:grad_probes}
\end{figure}

To test whether the gradient profile that motivates the RoSA schedule is specific to Llama-2-7B, we ran short dense fine-tuning probes on Mistral-v0.1-7B and Qwen-2.5-3B and recorded the average per-layer $\ell_2$ gradient norms. Figure~\ref{fig:grad_probes} shows the profiles. In both models, a single early layer shows an isolated spike, the following layers up to about Layer 11 are comparatively flat, and the middle-to-upper layers carry higher gradient norms, with the final layer remaining active. This is the same coarse structure as in Figure~\ref{fig:gradient_analysis}, which supports freezing a consecutive early prefix and rotating the memory budget through the middle-to-upper layers. The flat region extends further than in Llama-2-7B, so a longer frozen prefix may be possible for these models.

\section{Seed Variability and Stability Across Rotation Boundaries}
\label{app:stability}

\textbf{Seed variability.}
Table~\ref{tab:seeds} reports repeated runs with independent random seeds. On Llama-2-7B, a second run of the default RoSA configuration reaches 79.49\% average accuracy, compared with 79.61\% in the original run. On Qwen-2.5-3B, RoSA-DoRA reaches $81.83 \pm 0.32\%$ over three seeds, compared with 81.58\% for a single run of plain DoRA. The differences between seeds are small compared with the differences between methods in Section~\ref{sec:experiments}.

\begin{table}[h]
\centering
\small
\caption{Average commonsense reasoning accuracy (\%) across random seeds. For RoSA-DoRA we report the mean and standard deviation over three seeds.}
\label{tab:seeds}
\begin{tabular}{llc}
\toprule
\textbf{Setting} & \textbf{Run} & \textbf{Avg Acc.} \\
\midrule
Llama-2-7B, RoSA + NanoAdam & original run (Table~\ref{tab:detailed_results}) & 79.61 \\
 & seed 44 & 79.49 \\
\midrule
Qwen-2.5-3B, RoSA-DoRA & seed 42 (Table~\ref{tab:rosa_dora_qwen}) & 81.79 \\
 & seed 43 & 82.17 \\
 & seed 44 & 81.53 \\
 & mean $\pm$ std & $81.83 \pm 0.32$ \\
\midrule
Qwen-2.5-3B, DoRA & single run & 81.58 \\
\bottomrule
\end{tabular}
\end{table}

\begin{figure}[h]
\centering
\includegraphics[width=0.75\textwidth]{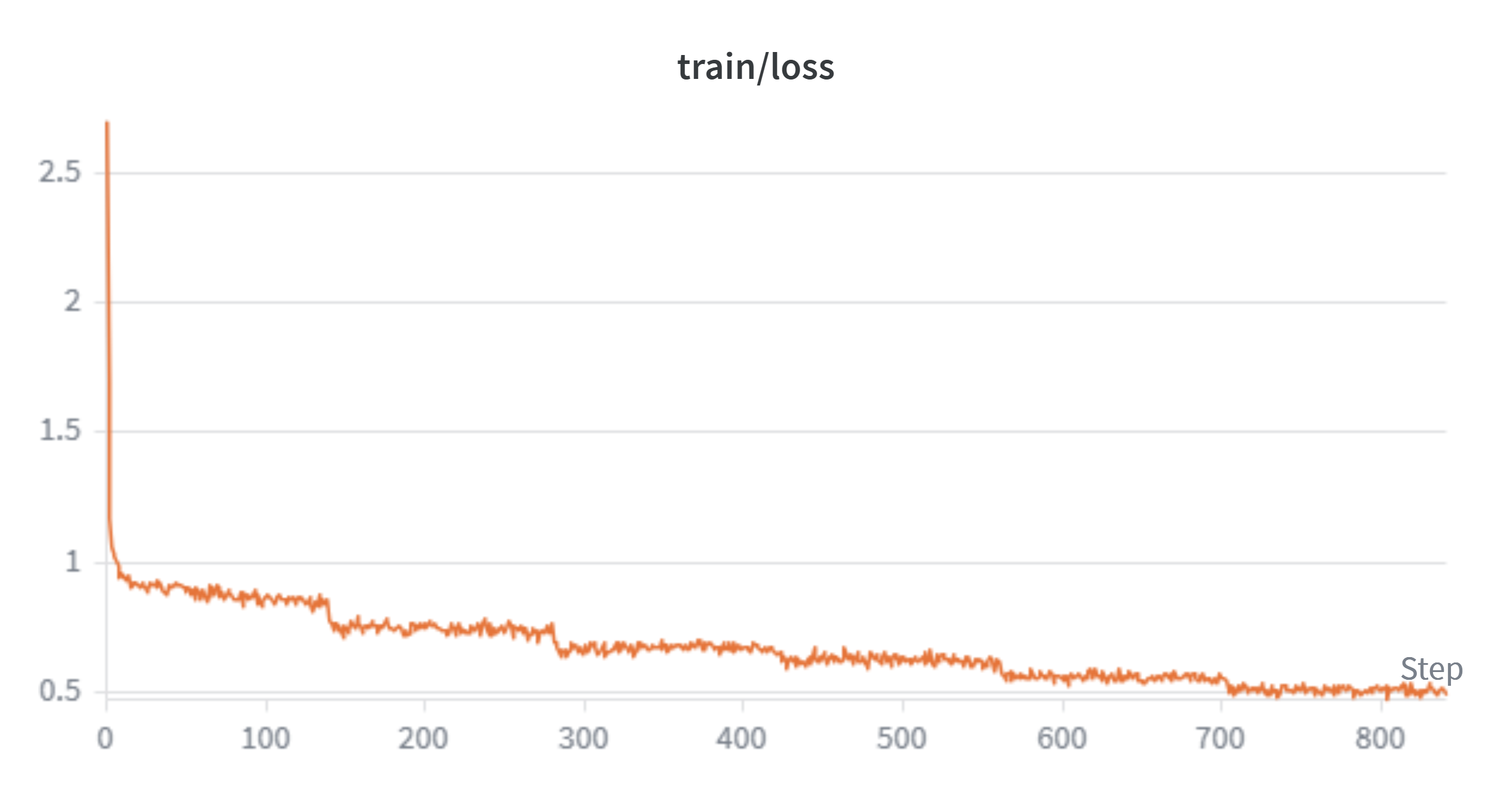}
\caption{Training loss of the default RoSA run on Llama-2-7B (two cycles, 6 epochs, 79.61\% average accuracy). Each step-down coincides with an epoch boundary, where the active group changes; the third marks the return to the first group at the start of the second cycle.}
\label{fig:training_loss}
\end{figure}

\textbf{Stability across rotation boundaries.}
Since RoSA re-initializes the optimizer states of a group whenever it becomes active (Section~\ref{sec:method}), one could expect loss spikes at rotation boundaries. We do not observe this in the default Llama-2-7B run (Figure~\ref{fig:training_loss}). During the first active group, the training loss drops from above 2.5 to about 0.85. It then steps down at every one of the five following group transitions, one per epoch, and ends at about 0.5. This includes the return to the first group at the start of the second cycle, where these layers are revisited with freshly initialized optimizer states. This indicates that the cold start does not destabilize training, and that layers adapted in earlier stages remain compatible with the representations produced by later updates when they are revisited.

\end{document}